\documentclass[conference,compsoc]{IEEEtran}
\IEEEoverridecommandlockouts
\usepackage{cite}
\usepackage{amsmath,amssymb,amsfonts}
\usepackage{algorithm}
\usepackage{algorithmic}
\usepackage{graphicx}
\usepackage[normalem]{ulem} 
\usepackage{textcomp}
\usepackage{xcolor}
\usepackage{url}
\def\BibTeX{{\rm B\kern-.05em{\sc i\kern-.025em b}\kern-.08em
    T\kern-.1667em\lower.7ex\hbox{E}\kern-.125emX}}
    
\newcommand{\rom}[1]{\uppercase\expandafter{\romannumeral #1\relax}} 
\newcommand{\D}{\boldsymbol{D}}
\newcommand{\hD}{\boldsymbol{\hat{D}}}

\newcommand{\T}{\boldsymbol{T}}

\newcommand{\ThD}{\boldsymbol{\T_{\hat{D}}}}

\newcommand{\B}{\boldsymbol{B}}

\newcommand{\vx}{\boldsymbol{x}}
\newcommand{\vy}{\boldsymbol{y}}
\newcommand{\vX}{\boldsymbol{X}}
\newcommand{\vY}{\boldsymbol{Y}}

\newcommand{\vd}{\boldsymbol{d}}

\newcommand{\vD}{\boldsymbol{D}}
\newcommand{\RNum}[1]{\uppercase\expandafter{\romannumeral #1\relax}}

\newcommand{\review}[1]{{\color{black}#1}}

\begin{document}

\title{Lightweight Conceptual Dictionary Learning \\for Text Classification Using  Information Compression}

\author{\IEEEauthorblockN{Li Wan$^\dag$, Tansu Alpcan$^\dag$,  Margreta Kuijper$^\dag$, and Emanuele Viterbo$^\ddag$}
\IEEEauthorblockA{$\dag$ Department of Electrical and Electronic Engineering, University of Melbourne, Parkville, VIC 3010, Australia. \\
$\ddag$ Department of Electrical and Computer Systems Eng., Monash University, Clayton, VIC 3800, Australia\\
Email: lwan3@student.unimelb.edu.au;\{tansu.alpcan, mkuijper\}@unimelb.edu.au; emanuele.viterbo@monash.edu}
\\
}

\maketitle

\begin{abstract}

We propose a novel, lightweight supervised dictionary learning framework for text classification based on data compression and representation. This two-phase algorithm initially employs the Lempel-Ziv-Welch (LZW) algorithm to construct a dictionary from text datasets, focusing on the conceptual significance of dictionary elements. Subsequently, dictionaries are refined considering label data, optimizing dictionary atoms to enhance discriminative power based on mutual information and class distribution. This process generates discriminative numerical representations, facilitating the training of simple classifiers such as SVMs and neural networks. We evaluate our algorithm's information-theoretic performance using information bottleneck principles and introduce the information plane area rank (IPAR) as a novel metric to quantify the information-theoretic performance. Tested on six benchmark text datasets, our algorithm competes closely with top models, especially in limited-vocabulary contexts, using significantly fewer parameters. \review{Our algorithm closely matches top-performing models, deviating by only ~2\% on limited-vocabulary datasets, using just 10\% of their parameters. However, it falls short on diverse-vocabulary datasets, likely due to the LZW algorithm's constraints with low-repetition data. This contrast highlights its efficiency and limitations across different dataset types.} 
\end{abstract}


\section{Introduction}

Text classification has been widely used in the real world including sentiment analysis \cite{li2020neural,fei2020topic}, review rating prediction \cite{seo2017interpretable}, topic categorization \cite{johnson2017deep,guo2019document}. Feature learning for text datasets is an important task in natural language processing (NLP) because most of the machine learning models require numerical inputs. Modern neural network-based methods use inherently black-box models, which provide limited conceptual meaning and can be computationally very demanding. This paper present a novel, lightweight supervised dictionary learning framework for learning text features, which is based on data compression and representation. This two-stage algorithm emphasises the conceptual meaning of dictionary elements in addition to classification performance. We use information-theoretic concepts such as
information bottleneck to investigate the performance of our algorithm with respect to the maximum achievable performance boundary on the information plane. Specifically,
we define a novel metric, information plane area rank (IPAR), to practically quantify the information-theoretic performance.

Combinations of word embedding and neural networks have achieved great success in many text classification tasks, including sentiment analysis, topic categorization, spam detection, and so on.
However, the neural network is still a black-box model with uninterpretable hidden layers in between. This problem is especially important in text datasets. Unlike the image datasets, where each attribute is a pixel that is not intuitive to humans, the basic attribute in text datasets is a word that has a straightforward conceptual meaning for us. The uninterpretable structures in neural networks prevent us from directly observing these. Moreover, modern NLP models can be very large and computationally demanding.

To address the above issues, this paper present a novel supervised conceptual dictionary learning algorithm using information compression. Some data compression algorithms such as Lempel-Ziv-Welch (LZW) explicitly map the text dataset into vector space with a generated dictionary. The dictionary is generated by finding the repeated substrings in the data, which is a white-box algorithm with an interpretable structure and preserves the conceptual meanings of words. The performance of the algorithm is largely based on the generated dictionary as all the data will be vectorized based on this dictionary. The challenge then is to find the optimal dictionary and representation for the given text dataset.  We adapt the original LZW algorithm into word-level LZW and letter-level LZW to generate dictionaries and representations for text datasets. Labels are also taken into consideration to achieve better classification performance. We update the dictionary generated from the LZW algorithm by choosing the atoms with higher discriminative power so the representation is also discriminative. This is an interpretable dictionary learning algorithm that can generate representations with conceptual meanings. 
In addition, the linear scanning strategy in the LZW algorithm requires fewer computation resources.

An important aspect of our algorithm is the use of information bottleneck concepts.
Tishby et al. \cite{tishby2000information} proposed information bottleneck that models the trade-off between compression and relevance for data representation. Information bottleneck formalizes this trade-off as an optimisation problem, which provides an information-theoretic boundary on the classification performance of a given compressed representation. In order to evaluate the performance of our algorithm, an information-theoretic analysis is presented using the information bottleneck approach. 

\begin{figure*}
    \centering
    \includegraphics[width = \linewidth]{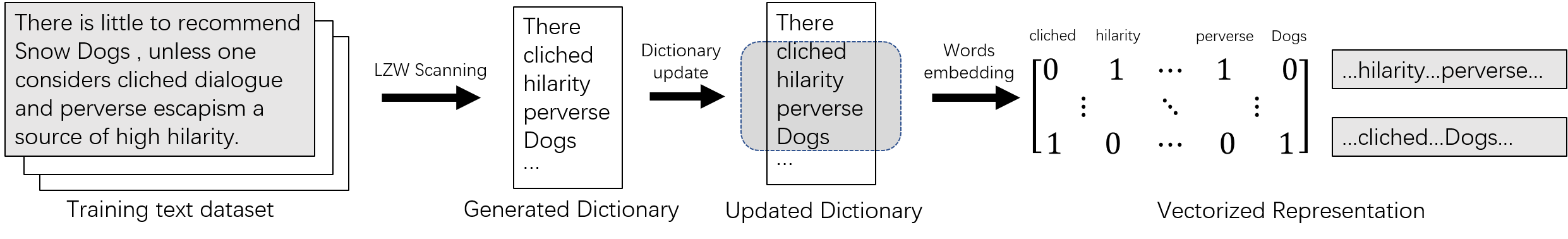}
    \caption{A flow chart of our algorithm. Given a text dataset, the LZW algorithm is implemented to generate a dictionary either at word-level or character-level. Then, the dictionary is updated by selecting a subset of atoms with high discriminative to better accomplish classification tasks. Labels are taken into consideration during the selection. With the updated dictionary, the text dataset is vectorized into a sparse vector representation. Finally, simple classifiers such as SVMs or neural networks can be trained for classification tasks.}
    \label{fig:flow_chart}
\end{figure*}

The main contributions of this paper are as follows:
\begin{enumerate}

\item A novel text data representation method based on classical data compression and discriminative power maximization. Our method generates a dictionary and vectorizes the dataset based on a classical data compression scheme, the Lempel-Ziv-Welch (LZW) algorithm. \review{The method does not require any prior knowledge of the dataset and is particularly suited for languages with structured elements, such as alphabetic systems exemplified by English.}

\item An algorithm for text classification with interpretable structures, and competitive performance, as depicted in Fig. \ref{fig:flow_chart}. Each step in the algorithm is a white-box step with interpretable meanings. For datasets with a lower richness of vocabulary, our algorithm outperforms state-of-the-art black-box algorithms.

\item Our algorithm is lightweight in terms of model size (number of parameters) in comparison to very large modern NLP models such as GPT \cite{radford2019language} or Bart \cite{lewis2019bart}.

\item An information-theoretic analysis of the framework from an information bottleneck \cite{tishby2000information} perspective. We also observe a shift of the optimal boundary in our dictionary learning cases. A novel information plane metric, information plane area rank (IPAR), is proposed to quantify the information-theoretic performance.

\item Experiments on 6 benchmark text classification datasets. We also provide insights into the properties of a dataset on which a data compression based method can work well on the information plane and the accuracy criteria. 
\end{enumerate}

This paper extends our previous work \cite{wan2019interpretable} with the following new points: 1) A comparison of our algorithm with state-of-the-art text classification algorithm on more benchmark datasets. 2) The observation of a potential shift of the optimal boundary in the theoretical analysis and our explanation. 3) Insights on when our algorithm performs well using the novel IPAR metric.

The organization of the paper is as follows. In Section \RNum{2}, we present the basic definitions and formulate the text classification problem. In Section \RNum{3}, the algorithm is proposed and the implementation details are presented. In Section \RNum{4}, an information-theoretic analytical method is presented. In Section \RNum{5}, our algorithm is implemented on six benchmark datasets. Experimental results and discussion are presented. Section \RNum{6} concludes the paper. In Section \RNum{7}, we show the related work of this paper regarding text classification, dictionary learning and information compression.

\section{Basic Definitions and Problem Formulation}

In this section, we adopt a probabilistic approach and establish a connection to supervised learning. A basic model of dictionary learning is also presented. These definitions and models will be revisited in the following sections to illustrate our algorithms and analytical tools. Table \ref{table:notation} is a notation table that describes all the variables in this paper.

\begin{table}[tbp]
    \centering
    \caption{Notation Table}
    \begin{tabular}{|c|c|}
    \hline
    Symbol & Meaning\\
    \hline
    $S = \{(\vx_i,\vy_i)\}_{i=1}^{N}$ & sample set\\
    $\mathcal{S}$ & distribution of sample set\\
    $N$ & number of samples\\
    $\vx$; $\vy$ & data sample; label\\
    $\mathcal{X}$; $\mathcal{Y}$ & data sample domain; label domain\\
    $\vX$; $\vY$ & data sample matrix; label matrix\\
    $d_0$; $d_1$ & data sample dimension; label dimension\\
    $X$; $Y$ & r.v. for data sample; label r.v.\\
    $x$; $y$ & random sample of the corresponding r.v. \\
    $\vD = \{\vd_j\}_{j=1}^M$ & dictionary matrix\\
    $M$ & number of atoms\\
    $\vd$ & dictionary atom\\
    $\boldsymbol{T}$ & sparse representation matrix\\
    $\boldsymbol{t}$ & sparse representation vector\\
    $T$ & sparse representation r.v.\\
    $t$ & a random sample of r.v. \\
    \hline
    \end{tabular}
    \label{table:notation}
\end{table}

\subsection{Probabilistic Model}

A typical supervised machine learning dataset $S = \{(\vx_i,\vy_i)\}_{i=1}^{N}$ is a collection of $N$ samples that are sampled from a distribution $\mathcal{S}$ over a domain $\mathcal{X} \times \mathcal{Y}$, where $\mathcal{X}$ is the instance domain and $\mathcal{Y}$ is the label domain. Normally, $\mathcal{X}$ is a subset of $\mathbb{R}^{d_0}$ and $\mathcal{Y}$ is a subset of $\mathbb{R}^{d_1}$, where $d_0$ is the instance dimension, and $d_1$ is the label dimension. We use the matrix $\vX\in\mathbb{R}^{N\times d_0}$ to represent a collection of the instances vectors $\vX = \{\vx_1;\vx_2;...;\vx_N\}$, and $\vY\in\mathbb{R}^{N\times d_1}$ to represent a collection of the label vectors $\vY = \{\vy_1;\vy_2;...;\vy_N\}$.
In order to investigate the statistical property of the dataset, a random variable $X$ is used to map the sample space $\mathcal{X}$ to a measurable space, and a random variable $Y$ is used to map the sample space $\mathcal{Y}$ to a measurable space.       

Given a dataset $S$, its statistical properties can be described by two random variables $X$ and $Y$. The entropy of $X$
\begin{equation}
    H(X) = \sum_{x\in X} p(x)\log\frac{1}{p(x)},
\end{equation}
measures the uncertainty of the random variable $X$. 

The mutual dependency of the instances and labels $X, Y$ can be measured by mutual information, which has the form

\begin{equation}
    I(X,Y) = H(X) - H(Y|X),
\end{equation}
where $H(Y|X)$ is the conditional entropy defined as
\begin{equation}
   H(Y|X) = \sum_{x \in X,y \in Y} p(x,y)\log\frac{1}{p(y|x)}.
\end{equation}

These information theory metrics can describe and quantify the probabilistic properties for a given dataset, which is of great significance in the analytical method.

\subsection{Text Classification Problem}

Text classification is one of the basic problems in natural language processing, which is commonly used in sentiment analysis, topic classification, spam filtering and so on. Given a text dataset $(X, Y)$, $X$ is usually a set of corpora and $Y$ represents the corresponding labels. A text classification algorithm takes the corpus as input, analyses the content of input and predicts the label as output.

The classification process can be further broken down as most machine learning algorithms (especially neural networks) require numerical inputs. Therefore, a tokenization process that transforms the corpus into numerical vectors is essential in text classification tasks. As for the results, in addition to classification accuracy, the conceptual meaning of one or several words in the decision-making process should also be investigated. Lastly, a lightweight algorithm with competitive performance is always preferred as it requires less computational resources and suits all kinds of situations. The problem we are solving in this paper is to propose a lightweight text classification algorithm with competitive performance and conceptual words in the decision-making process. Information analyses based on probabilistic models are used to further investigate the algorithm.

\section{Related Work}
In this section, we recall some text classification models including CNN, RNN and self-attention. Connections between dictionary learning and our algorithm are introduced since our algorithm is a dictionary-based algorithm. Information compression in machine learning, especially text classification, is presented to support the theoretical analysis in our work.

\subsection{Text Modeling and Classification}
Text classification algorithms are typically categorized into statistical, machine learning (ML), and deep learning approaches, with the latter often being favored in supervised tasks for their superior accuracy. Central to these methods is the vectorization of text data, a process that transforms text into numerical vectors, also known as word embeddings. This transformation, however, poses challenges due to the variable length of texts and the complexity of word contexts. Traditional methods like Bag-of-words (BoW), n-gram, and One-hot encoding, though widely used, can lead to sparse, high-dimensional vectors not ideal for deep learning models \cite{goldberg2017neural} and may overlook semantic relationships between words.

Recent strides in text classification have been made with advanced word embedding techniques such as Word2Vec \cite{mikolov2013efficient} and GloVe \cite{pennington2014glove}, and the application of sophisticated neural network architectures like CNNs \cite{zhang2015sensitivity}, LSTM networks \cite{ruan2017sequential}, and pre-trained models including GPT-2 \cite{radford2019language} and BERT \cite{lewis2019bart}. These models, although computationally intensive, excel in capturing local semantic correlations and maintaining contextual information, marking significant progress in natural language processing (NLP) tasks.

\subsection{Dictionary Learning}
Dictionary learning and sparse representation (DLSR) have recently gained traction in the machine learning domain. DLSR algorithms, which can be supervised \cite{mairal2009supervised,mairal2011task} or unsupervised \cite{wright2008robust}, have shown improved classification performance when the dictionary learning is integrated with classifiers \cite{akhtar2016discriminative,gangeh2013kernelized}. These algorithms leverage various mathematical concepts, such as the information bottleneck \cite{fulkerson2008localizing}, Fisher discrimination criterion \cite{yang2011fisher}, and histograms of dictionary atoms \cite{gangeh2011dictionary}, to enhance the learning process.

The advent of hierarchical and deep dictionary learning algorithms, inspired by the structure of neural networks, marks a significant evolution in the field \cite{tariyal2016deep,mahdizadehaghdam2019deep,9112646}. These advanced models employ multiple layers and utilize backpropagation or the chain rule to learn various dictionaries and their representations, offering a sophisticated approach to tackle high-dimensional datasets. The performance of these algorithms, particularly in image classification, is on par with state-of-the-art convolutional neural networks.

Moreover, the interpretability of dictionary learning, a vital aspect in applications like fMRI \cite{mensch2016dictionary}, spatio-temporal analysis \cite{kim2010sparse}, and face recognition \cite{liu2016face}, stems from the clear relationship between the original dataset and its sparse representation, as well as the visibility of dictionary atoms. The optimization process in dictionary learning, involving reconstruction error, sparsity of the coefficients, and mutual information, among others, provides a controllable and interpretable framework. This makes DLSR a robust approach for feature generation and classification tasks in machine learning.

\subsection{Information Compression and Text Classification}
\review{Information theory is pivotal in analyzing communication systems and signal processing, covering crucial aspects like compression, transmission, and channel capacity. Data compression algorithms play a significant role in transforming text datasets into feature vectors \cite{sculley2006compression}, thereby bridging machine learning and compression. This paper sheds light on various source coding algorithms and introduces metrics integral to machine learning, fostering a blend of innovation and practicality.}

\review{The application of data compression for string character classification dates back to the early 2000s. Methods estimating cross-entropy between text corpora and labels have been instrumental in topic classification \cite{benedetto2002language, frank2000text}. This research offers two main strategies to enhance text classification via data compression: enhancing the compressor itself or refining the distance metric. Building on this, \cite{coutinho2015text} introduced normalized information distance (NID) to improve classification accuracy. More recent works like \cite{kasturi2022text} leverage advanced compressors for more effective data representation, boosting the accuracy further.}

\review{However, the intricate balance between data compression and precise representation, particularly in the context of learning procedures, remains a less explored topic. This is where the concept of the information bottleneck becomes relevant \cite{tishby2000information, tishby2015deep}. Initially conceptualized to elucidate information flow in neural networks, the information bottleneck principle aptly aligns with our discussion, integrating information compression into the learning process for text classification tasks. It offers a theoretical optimal boundary for balancing information compression and preservation. However, research examining the position of compression algorithms relative to this optimal boundary is scant, indicating a significant scope for exploration and development in this area.}

\section{Supervised Dictionary Learning Using LZW}

\subsection{Data Compression and Dictionary Learning}

Lempel-Ziv-Welch (LZW) is a universal lossless data compression algorithm that was proposed in \cite{welch1984technique} as an improvement of the LZ77 and LZ78  algorithms proposed in \cite{ziv1977universal,ziv1978compression}. 
Given a sequence, the LZW algorithm firstly initializes the dictionary with all the strings of length one in the sequence and then expands it by iteratively adding the longest matching string in the dictionary followed by the next symbol in the input sequence, until the end of the sequence.

A dictionary $\vD$ is a matrix that contains various atoms $\boldsymbol{d}$ that can be either variable length or fixed length. Given a dataset $S$, if we quantize (binarize) and sequentialize the dataset $S$ into a sequence, the scanning process in the Lempel-Ziv-Welch algorithm can generate a dictionary $\vD = \{\vd_j\}_{j=1}^M$ where each atom $\boldsymbol{d_j}$ is a binary sequence, which is a subset of the original sequence and the length of $\boldsymbol{d_j}$ varies from one to a pre-defined value.

Suppose that the generated dictionary $\vD$ is closely related to the given dataset $S$ such that each sample $\vx_i$ can be represented by one or several atoms in the dictionary. A column vector $\boldsymbol{t_i}\in\mathbb{R}^M$ describes whether each atom is used or not in this representation. The matrix $\boldsymbol{T_{\vD}} = [\boldsymbol{t_1},\boldsymbol{t_2},\dots,\boldsymbol{t_N}]^T$ is a representation matrix that describes how the dataset $S$ can be represented by the dictionary $\vD$. Note that this representation matrix can be computed by solving different types of optimization problems. A novel optimization method to compute $\boldsymbol{T_{\vD}}$ called minimum-distortion-longest-match (MDLM) will be introduced in section V. In this section, we just assume that a representation is computed so that the reconstructed sample can be described as 
\begin{equation}
    \boldsymbol{\hat{s}}_i = \sum_{j=1}^{M} t_{i,j}\cdot \boldsymbol{d}_j \cdot 2^{\sum_{k=j+1}^M |\boldsymbol{d}_k|\cdot t_{i,k}},
\end{equation}
where $t_{i,j}\in \{0,1\}$ is $j$-th element of the vector $\boldsymbol{t_i}$ that indicates whether the atom $\boldsymbol{d}_j$ is used in the reconstruction, and $|\boldsymbol{d}_j|$ is the length of binary atom  $\boldsymbol{d_j}$. The exponent acts as a shifter here so the concatenation is achieved. For example, assuming that $\boldsymbol{t_1} = [1,0,0,1]$ and $\boldsymbol{D} = \{1101,0010,1111,0000\}$, then $\boldsymbol{\hat{s}}_1 = 11010000$. Note that this binary representation can be generalized to letters for text datasets or decimal numbers for image datasets. The reconstructed sample is not necessarily identical to the original one for two reasons: 1) some atoms in the dictionary will be discarded during the dictionary updating process; 2) the dictionary may not be able to losslessly reconstruct a sample that it has never seen before.

Similarly, the random variable $T$ is used to describe the statistical properties of a general representation $\boldsymbol{T}$, where the subscript can be used to indicate which dictionary the representation relies on.

\subsection{Dictionary Generation}
The way our algorithm creates the dictionary is similar to the LZW algorithm except that the mapping between dictionary atoms and binary bits is omitted while the scanning step is preserved. The main idea of the scanning process is to find the longest substring that is already in the dictionary and add the next string to the dictionary with the next code. While the original LZW algorithm scans the sequence on a bit/character level, we extend the algorithm to a word level. Specifically, instead of finding the long substrings, the longest n-gram is found and added to the dictionary with the next available word. This adaptation is made to achieve better classification performances for the datasets with a high richness of vocabulary and save implementation time. Both versions of the algorithms (character-version and word-version) will be discussed and implemented in the rest of our work.

\review{In our method, given a dataset $\boldsymbol{S}$, we concatenate its data samples to form a serialized sequence. This sequence is then processed using the modified LZW scanning technique, which generates a dictionary $\boldsymbol{D}$. It’s important to note that during this process, some dictionary items may span across the boundaries of adjacent data samples. Typically, such cross-sample items are considered irrelevant, particularly in text datasets, because they do not correspond to meaningful words within any individual data sample.

Instead of inserting a stop symbol at the end of each sample to avoid these cross-sample items,  we preserve them for two primary reasons: firstly, although generally considered semantically meaningless, these cross-sample items could potentially hold value in certain datasets. Secondly, even if these items are generated, they are likely to be filtered out in the subsequent dictionary updating process, thereby mitigating the concern of retaining irrelevant data.}

 \begin{algorithm}[h]
  \caption{Dictionary \textbf{update} based on discriminative power maximization (DMP)}\label{euclid}
  \begin{algorithmic}[1]
    \STATE \textbf{Input:} LZW dictionary $\boldsymbol{D}$, dataset $S$, labels $\boldsymbol{L}$, expected dictionary size $k = |\boldsymbol{\hat{D}}|$ 
    \STATE \textbf{Initialization:} $\hD = \emptyset$ to store the selected atoms.
     \FOR{every atom $\boldsymbol{d_i}$ in $\boldsymbol{D}$}
        \STATE Compute occurrence rate $p_{\boldsymbol{d}}$
        \STATE Compute discriminative \textit{dispower}$(\boldsymbol{d})$
    \ENDFOR
    \STATE Select $k$ atoms in $\boldsymbol{D}$ with highest discriminative power to form $\boldsymbol{\hat{D}}$
    \STATE \textbf{Output}: dictionary $\hD$
  \end{algorithmic}
\end{algorithm}

\begin{algorithm}[h]
\caption{Minimum-distortion-longest-match (MDLM) Algorithm}
\begin{algorithmic}[1]

    \STATE \textbf{Input:} LZW dictionary $\hD$, dataset $S$
    \STATE \textbf{Initialization:} $\ThD$ is a zero matrix with size $|\B|\times |\D|$.
    \FOR{every $\boldsymbol{s_i}$ in $S$}
     \STATE $temp = s_i$
     \WHILE{length(temp) $\neq$ 0}
     \FOR{every atom $\boldsymbol{d_j}$ in $\boldsymbol{D}$}
        \STATE $l = \text{length}(d_j)$ 
        \STATE $dis[j] = \text{Hamming Distance}(\boldsymbol{s_i}[0:0+l], \boldsymbol{d_j}) / l$
    \ENDFOR
    \STATE $\hat{j}$ is the index of the smallest element in $dis$. If there are multiple elements have the same average distance, we choose the one with longest length.
    \STATE $\ThD[i][j] = 1$
    \STATE $temp = temp\big[dis[\hat{j}]:\,]\big]$
    \ENDWHILE
    \ENDFOR
    \STATE \textbf{Output}: sparse representation matrix $\ThD$
\end{algorithmic}
\end{algorithm}
\subsection{Dictionary Update}
We are motivated to further update the dictionary for the following reasons: 1) The dictionary $\boldsymbol{D}$ is created regardless of the labels, which leads to poor performance in classification tasks. Therefore, the dictionary should be updated by taking labels into consideration to better accomplish the classification tasks. 2) The LZW dictionary is only lossless for the training set, while information loss happens when new test date samples are passed through the system. Since information loss is inevitable, how to control the information loss and how to compress is what we focus on in this paper. The information bottleneck method provides a way to preserve the information that is relevant to the label while information loss occurs. 3) \review{Previous studies \cite{tishby2015deep, gabrie2018entropy, lee2021reducing, goldfeld2019estimating, tang2019markov} have observed that information compression can enhance classification performance. In line with these findings, our approach seeks to eliminate redundant atoms, aiming to achieve a balance between lightweight model architecture and improved accuracy.}

A discriminative-power-maximization (DPM) method to select a subset atoms from $\boldsymbol{D}$ to obtain $\boldsymbol{\hat{D}}$ and its detailed explanation to measure the discriminative power are presented next.
Such a discriminative dictionary can promote the samples from different classes described by different atoms, which leads to a better performance in the classification tasks. The discriminative power of each atom in the dictionary is measured by the distribution of its occurrence in different classes of samples. Atoms that occur more in the samples from the same class will increase its discriminative power while the atoms that have an even spread over different classes will have small discriminative power. A certain amount of atoms with the highest discriminative power will make up $\hD$. Given an atom $\boldsymbol{d}$ and a collection of samples and labels $S = \{(\boldsymbol{s}_i,l_i)\}_{i=1}^N$ with $M$ different classes, the occurrence rate of $\boldsymbol{d}$ for class $m$ is computed as 
 \begin{equation}
     p^m_{\boldsymbol{d}} = \frac{c^m(\boldsymbol{d})}{\sum_{m=1}^M c^m(\boldsymbol{d})},
 \end{equation}
 where $c^m(\boldsymbol{d})$ counts the number of samples with label $m$ including atom $\boldsymbol{d}$ as a subset. The discriminative power is measured as
 \begin{equation}
     \textit{dispower} (\boldsymbol{d}) = \sum_{m=1}^M c^m(\boldsymbol{d})\cdot\bigg[\log_2(M) - \sum_{k=1} ^ M  p^k_{\boldsymbol{d}} \cdot \log_2 \frac{1}{p^k_{\boldsymbol{d}}}\bigg],
     \label{dp}
\end{equation}
 which is the Kullback–Leibler (KL) divergence between the uniform distribution and the distribution of the atoms, multiplied by the occurrence of the atom.
 The intuitive explanation is that an atom is likely to have higher discriminative power when it only appears in a certain class of samples, which means its distribution regarding to the labels is not uniform. In addition, the number of occurrences is also considered because an atom that appears multiple times has higher discriminative power compared with an atom that only appears once. A simple example is given to demonstrate how discriminative power is computed. Assuming we have 6 samples $\{\boldsymbol{s}_1,\boldsymbol{s}_2,\boldsymbol{s}_3,\boldsymbol{s}_4,\boldsymbol{s}_5,\boldsymbol{s}_6\}$ with labels $\{1,1,2,2,3,3\}$, atom $\boldsymbol{d}_1$ is $\{\boldsymbol{s}_1,\boldsymbol{s}_2,\boldsymbol{s}_4,\boldsymbol{s}_6\}$ which is a subset of the samples list, then its occurrence rate is $\{0.25,0.5,0.25\}$ and its discriminative power is 2. Given another atom $\boldsymbol{d_2}$ as $\{\boldsymbol{s}_1, \boldsymbol{s}_2\}$, its occurrence rate is $\{1,0,0\}$ and its discriminative power is 4. Therefore, $\boldsymbol{d}_2$ is an atom with higher discriminative power, which can be explained intuitively as it creates no ambiguity to classify a sample including $\boldsymbol{d}_2$ into class 1. The detailed implementation is shown in Algorithm 2.

\subsection{Sparse Coding}
A lossless reconstruction (decoding) procedure in LZW is achieved by a mapping between the dictionary atoms and their positions in the encoded sequence. However, with the updated dictionary, a lossy compression is required here for two reasons: 1) The dictionary is updated so a lossless reconstruction based on the updated dictionary is not guaranteed; 2) For classification tasks, samples from the testing set cannot be reconstructed losslessly by the atoms in the dictionary. Therefore, Algorithm 2 is presented as a lossy reconstruction algorithm to address this problem.

Given the dictionary $\hD$ and the original dataset $S$, finding the spare representation $\boldsymbol{T}$ is can be interpreted as a set cover problem proved to be NP-hard\cite{korte2012combinatorial}. To find an approximate solution, we propose the minimum-distortion-longest-match (MDLM) algorithm to find the sparse representation $\ThD$. The MDLM algorithm (see Algorithm 2) considers the sparsity and the reconstruction error at the same time by always choosing the atom with the minimum distortion and the longest length to represent the sample. We use the hamming distance divided by the length of the atom to measure the distortion between an atom and a sample. The time complexity of MDLM is $O(|\boldsymbol{s}| \cdot |\hD|)$, where $|\boldsymbol{s}|$ is the length of each sample in dataset $S$ and $|\hD|$ is the number of atoms in dictionary $\hD$.

\review{This approach is indeed somewhat analogous to TF-IDF \cite{SaltonMcGill1983} in leveraging the occurrence frequency
of an atom to generate a sparse representation of a corpus. However, a key distinction lies in the nature of the algorithms: TF-IDF is an unsupervised metric that does not involve class labels, while our method is supervised and incorporates class labels
during the computation process. In our algorithm, the class labels play a crucial role in determining the
discriminative power of an atom, influencing its selection or rejection during the dictionary construction
phase. This integration of class labels is a fundamental aspect that differentiates our method from
traditional TF-IDF.}

\section{An information-theoretic Analysis}
In this section, a novel analysis of the compression and representation model of our dictionary learning approach is presented. 


\subsection{Analysis using Information Bottleneck}
Information bottleneck studies the trade-off between compression and representation \cite{tishby2000information}. The information bottleneck method requires the joint probability $p(X,Y)$ to compute the theoretical compression and representation boundary for any learning algorithm. The boundary is generated by solving the multi-objective optimization problem (\ref{IB}) as the weighting parameter $\beta$ increases from 0 to a large number.

\begin{equation}
    l_{IB} = \min_T I(X;T) - \beta I(Y;T),
    \label{IB}
\end{equation}
where $I(X;T)$ measures the compression between a representation $T$ and the original dataset. $I(Y;T)$ measures the relevant information that a representation $T$ related to labels. Here, $\beta$ is a variable that weights the importance between compression and representation.

\begin{figure}[bt]
    \centering
    \includegraphics[width = 0.7\linewidth]{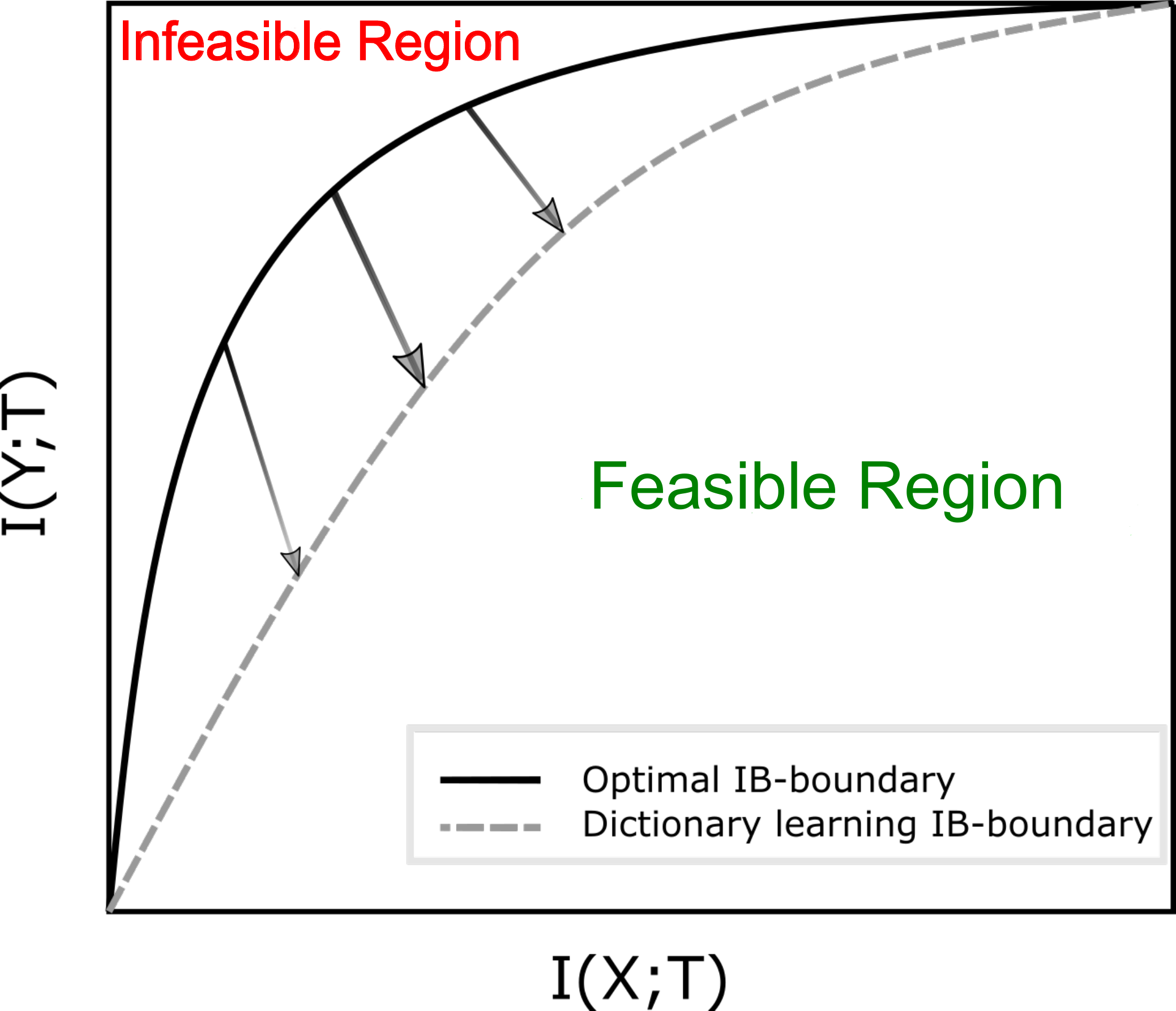}
    \caption{Information plane with feasible and infeasible regions of machine learning algorithms and Information Bottleneck (IB) boundary. The dashed line denotes a potential path of a dictionary learning algorithm. Arrows indicate the potential shift between the optimal information bottleneck boundary and the dictionary learning boundary.}
    \label{fig:ibplane}
\end{figure}

\review{In practice, the optimal boundary is usually not found by solving the optimization problem due to the curse of dimensionality \cite{kolchinsky2019nonlinear, belghazi2018mine}. In this work, we adopt the method in \cite{kolchinsky2019nonlinear} to estimate the optimal boundary.} The boundary divides the information plane, defined by a coordinate where $I(X;T)$ is the x-axis and $I(Y;T)$ is the y-axis, into two regions. The region over the boundary is the unachievable region while the region under the boundary is the achievable region. The boundary describes the best representation that an algorithm reaches when the compression rate is fixed. Fig. \ref{fig:ibplane} depicts the information plane created by information bottleneck. All the algorithms should be in the achievable region.

Given a dataset $S$ with the corresponding probabilistic properties represented by two random variables $X$, $Y$, its representation based on a dictionary $\hat{\vD}$ is defined as $T_{\hat{\vD}}$.  For any dictionary $\hD$, its position on the information plane is $\big(I(X;T_{\hat{\vD}}), I(Y;T_{\hat{\vD}})\big)$. As the size of $\hD$ increases, a possible trajectory on the information plane will be generated as shown by Fig. \ref{fig:ibplane}. The first item $I(X; T_{\hat{\vD}})$ is calculated as follows:

\begin{equation}
\begin{split}
    I(X; T_{\hat{\vD}}) &= H(X) - H(X|T_{\hat{\vD}})\\
    &= H(X) - \sum_{t \in T_{\hat{\vD}}} p(t) \sum_{x\in X} p(x|t)\log\frac{1}{p(x|t)},
\end{split}
\label{compression}
\end{equation}
which involves the estimation of probability density functions $p(X)$, $p(T_{\hat{\vD}})$ and $p(X|T_{\hat{\vD}})$. We can use the kernel density estimator \cite{bishop2006pattern} to estimate $p(X)$ and $p(X|T_{\hat{\vD}})$ while $p(T_{\hat{\vD}})$ is estimated by counting and dividing because $T_{\hat{\vD}}$ is a sparse matrix with only 0's and 1's. 

The representative mutual information $I(Y;T_{\hat{\vD}})$ is evaluated as follows:
\begin{equation}
\begin{split}
    I(Y; T_{\hat{\vD}}) &= H(Y) - H(Y|T_{\hat{\vD}})\\
    & = H(Y) - \sum_{t} p(t) \sum_{y} p(y|t)\log\frac{1}{p(y|t)},
\end{split}
\end{equation}
where $p(Y)$ is already estimated in the joint probability estimation $p(X,Y)$, $p(t)$ is estimated in (\ref{compression}) and $p(y|t)$ is estimated by counting the occurrence of the corresponding samples.

Problem described in (\ref{IB}) is a general form that provides an ideal boundary only based on the probabilistic properties of the dataset, regardless of the machine learning algorithm. In dictionary learning, additional constraints may apply as we either want to minimise the reconstruction error or have sparsity restrictions on the representation coefficients. Therefore, given a dictionary learning algorithm with a set of constraints described by the set $C$, the specific information bottleneck boundary should be
\begin{equation}
        \min_{p(t|x)} I(X;T) - \beta I(Y;T)
        \quad \text{s.t.} \quad T \in C \subset \mathbb{R}^{n_t},
    \label{con_opt}
\end{equation}
which is the constrained version of the  optimisation in ($\ref{IB}$). 

Comparing \eqref{con_opt} and \eqref{IB}, we observe
the following:
For the optimization problems described in (\ref{IB}) and (\ref{con_opt}) with a given $\beta$, $I(X;T'(\beta)) \geq I(X;T(\beta))$ and $I(Y; T'(\beta))\leq I(Y; T(\beta))$ where $T(\beta)$ is optimal solution for (\ref{IB}) and $T'(\beta)$ is the optimal solution for (\ref{con_opt}).

To see this, let $\Phi(\beta)$ be function of the optimal value of objective function (\ref{IB}) respect to $\beta$ with an optimal solution $T(\beta)$. Similarly we have $\Phi'(\beta)$, $T'(\beta)$ for (\ref{con_opt}). Since (\ref{con_opt}) is the constrained version of (\ref{IB}) with $C \subset \mathbb{R}^{n_t}$,  we have 
\begin{equation}
\begin{split}
    \Phi'(\beta) &= I(X;T'(\beta)) -\beta I(Y;T'(\beta))\\ 
    &\geq  I(X;T(\beta)) - \beta I(Y;T(\beta)) = \Phi(\beta).
\end{split}
\label{Inequality}
\end{equation}
Thus, for a given $\beta$, we either have $I(X;T'(\beta)) \geq I(X;T(\beta))$ or $I(Y;T'(\beta)) \leq I(Y;T(\beta))$, together with the fact that dictionary learning cannot do better than the original boundary, we end up with a potential shrinkage of the feasible region as depicted in Fig. \ref{fig:ibplane}. 
An intuitive explanation of this observation is that the constraints introduced by the dictionary learning algorithm may degrade the information-theoretic performance on the information bottleneck plane. In other words,
minimizing the reconstruction error in dictionary learning algorithms may potentially shift the optimal boundary computed by information bottleneck, resulting in 
shrinkage of the feasible region (Fig.~\ref{fig:ibplane}).

\subsection{Information Plane Area Ratio (IPAR)}

Each point on the information plane represents a compression-relevance pair, quantified by mutual information, which illustrates the informational performance of a given representation. The feasible region on this plane includes a set of achievable algorithms, each with its specific information performance profile. 

\begin{figure}[bt]
    \centering
    \includegraphics[width = 0.7\linewidth]{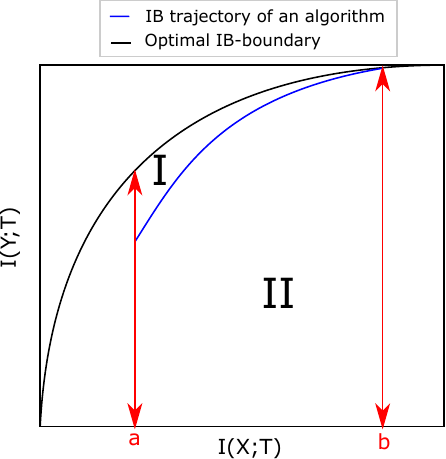}
    \caption{Information plane with an optimal IB-boundary and an example IB trajectory. The information trajectory is located below the optimal boundary, which divides the feasible region into two (region I and region II) if two vertical lines are added at the start ($I(X;T) = a$) and the end ($I(X;T) = b$) of the trajectory. The ratio between the area of region I and region II is defined as Information Plane Area Ratio (IPAR).}
    \label{fig:IPAR}
\end{figure}

The trajectory of any algorithm on the Information Bottleneck (IB) plane is constrained to lie below the optimal boundary. The trajectory’s start point ($a$) represents the maximum achievable compression by the algorithm, while its endpoint ($b$) denotes the minimum compression. These two points partition the feasible region into three distinct areas: Region I, Region II, and the remainder, as demonstrated in Fig. \ref{fig:IPAR}. 

The Information Plane Area Ratio (IPAR) is defined as the ratio between the areas of Region I and Region II. This metric is used to evaluate the information plane performance of a given algorithm. Assuming the optimal IB-boundary is represented by the function $f_{\text{IB}}(x)$, and the IB trajectory of an algorithm is denoted by $g_{\text{Alg}}(x)$, IPAR is mathematically expressed as:

\begin{equation}
\begin{split}
    \text{IPAR} & \equiv \frac{A(\text{Region I})}{A(\text{Region II})} \\ 
    & = \int_a^b \frac{f_{\text{IB}}(x) - g_{\text{Alg}}(x)}{g_{\text{Alg}}(x)} dx,
\end{split}
\end{equation}
where $A(\cdot)$ indicates the area of the specified region.

Region I represents the set of algorithms that outperform the given algorithm by achieving higher relevance (measured by $I(Y;T)$) at the same compression rate (measured by $I(X;T)$). In contrast, Region II embodies the set of algorithms that underperform. Therefore, the IPAR quantifies the balance between these regions, with a smaller ratio indicating a performance closer to the optimal on the information plane. \review{Check Algorithm \ref{alg:ipar} for a step-by-step guide on how to compute IPAR for a given dataset and a compression scheme.}

\begin{algorithm}[h]
  \caption{Calculation of Information Plane Area Ratio (IPAR)}\label{alg:ipar_conceptual}
  \begin{algorithmic}[1]
    \STATE \textbf{Input:} Dataset $S$, Compression algorithm $\mathcal{C}$ 
    \STATE \textbf{Procedure:}
    \STATE Compute the optimal Information Bottleneck (IB) boundary $f_{\text{IB}}(x)$ for dataset $S$.
    \STATE Obtain the information trajectory $g_{\mathcal{C}}(x)$ using compression scheme $\mathcal{C}$ and dataset $S$.
    \STATE Calculate the areas of Region I and Region II under the IB boundary and the information trajectory.
    \STATE Compute IPAR $\gets$ Area of Region I / Area of Region II
    \STATE \textbf{return} IPAR
    \STATE \textbf{Output:} IPAR: Information Plane Area Ratio
  \end{algorithmic}
  \label{alg:ipar}
\end{algorithm}

\begin{figure*}
    \centering
    \includegraphics[width = \linewidth]{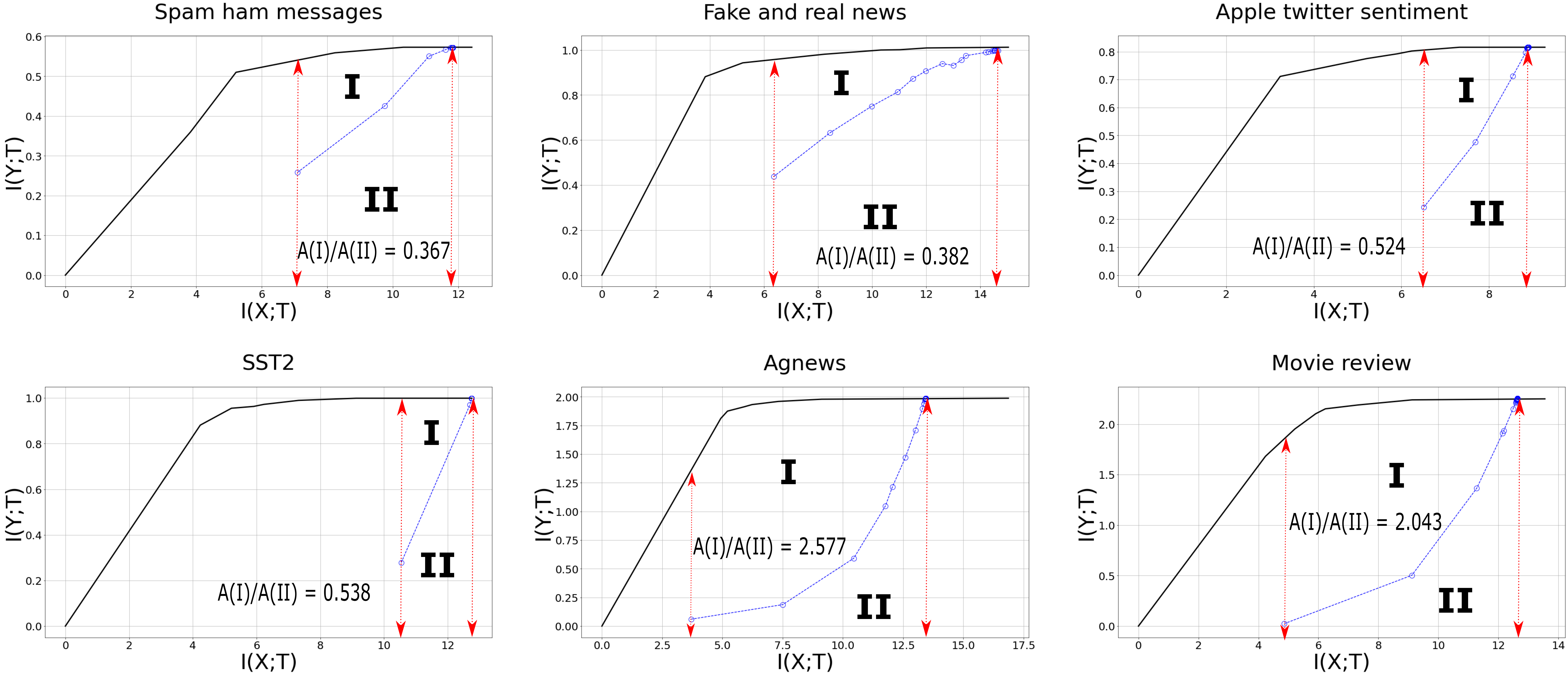}
    \caption{Information-theoretic analysis of the performance of our algorithm on six datasets. The black curve is the optimal information bottleneck boundary. The blue curve is the information trajectory when the number of atoms in the updated dictionary increases from 2 to a large enough value (no change is observed if the number is increased after a threshold). The dictionary is generated from the character-version implementation. The IPAR score for each dataset is 0.367, 0.382, 0.524, 0.538, 2.577, 2.043, as shown in the figures.}
    \label{fig:info_result}
\end{figure*}

\begin{table*}[h]
    \centering
    \caption{COMPARISONS OF TEST ACCURACY (\%) ON SIX BENCHMARK DATASETS}
    \begin{tabular}{|c||c|c||c|c||c|c|}
    \hline
         Algorithms & Spam ham messages & Fake and real news & Apple Twitter sentiment &  SST-2 & Agnews & Movie review  \\
         \hline
         \hline
         CNN-non-static \cite{zhang2010discriminative} & 97.18 & 96.12& 91.71 &  85.22 & 88.81 & 80.06\\
         \hline
         C-LSTM \cite{zhou2015c} & 97.43 & 95.15 & 92.78 &  87.80 & 86.73 & 78.41 \\
         \hline
         Bi-LSTM \cite{jang2020bi} & 98.14 & 95.17 & 93.42  & 87.74 & 91.62 & 79.71 \\
         \hline
         MC-CNN \cite{zhang2015sensitivity} & 97.92 & 96.78 & 90.72  & 88.13 & 92.11 & 81.10 \\
         \hline
         RCNN \cite{lai2015recurrent} & 96.11 & 97.42 & 90.13 &  87.42 & 91.54 & 82.13 \\
         \hline
         Self-Attentive \cite{lin2017structured} & 98.27 & 97.43 & 91.24  & 83.42 & 84.13 & 80.10 \\
         \hline
         SVM \cite{socher2013recursive} & 98.79 & 94.92 & 77.38 & 79.94 & 85.68 & 68.93 \\
         \hline
         fastText \cite{joulin2016bag} & 97.16 & 96.93 & 90.12  & 86.19 & 92.53 & 79.14 \\
         \hline
         Text GCN \cite{yao2019graph} & 96.52 & 97.89 & 92.48  & 90.68  & 90.62  & 76.74 \\
         \hline
         \textbf{Ours (words)} & 97.96 & 98.92& 83.48  & 81.52 & 56.41 & 43.61  \\
         \hline
         \textbf{Ours (characters)} & 98.02 & 98.76 & 81.24  & 83.57 & 54.21 & 29.41 \\
    \hline
    \end{tabular}
    \label{tab:result}
\end{table*}
\section{Experimental Evaluation}
In this section, we present the descriptions of different datasets, the experimental setup, the models we compared with, and the experimental results. 

\subsection{Datasets}
We implement our algorithm on 6 datasets. The first 4 datasets are  spam and ham messages 
\footnote{\url{https://www.dt.fee.unicamp.br/~tiago/smsspamcollection/}}\cite{cormack2007spam}, fake and real news\footnote{\url{https://www.kaggle.com/clmentbisaillon/fake-and-real-news-dataset}}\cite{ahmed2018detecting}, apple twitter sentiment 
\footnote{\url{https://data.world/crowdflower/apple-twitter-sentiment}} and SST2\footnote{\url{https://deepai.org/dataset/stanford-sentiment-treebank}} \cite{SocherEtAl2013:RNTN}. Each sample (a paragraph with one or multiple sentences) in these datasets has a short length (no more than 20 words) and has more repeated words. In contrast, the last two datasets are Agnews\footnote{\url{http://groups.di.unipi.it/~gulli/AG_corpus_of_news_articles.html}}\cite{zhang2015character}, and movie reviews\footnote{\url{https://www.cs.cornell.edu/people/pabo/movie-review-data/}} \cite{pang2005seeing}. These datasets are more complicated compared with the previous 6 with longer corpora and richer vocabulary.

\subsection{Experimental Setup}
\review{During the LZW scanning, all the texts in the training set are concatenated into one string for the scanning process. To ensure the results are meaningful, we strictly separated our dataset into training and testing sets. For datasets without a predefined split (e.g., Spam Ham Messages, Apple Twitter Sentiment), we allocated 20\% of the data as the testing set. For datasets with an existing split, we adhered to the provided partitioning. We implement our algorithm on the same dataset 10 times with shuffled concatenation order and no statistically significant difference in results is noticed. As for the classifier, we use linear support vector machines (SVMs) or one-hidden layer perception neural networks, whichever performs better.  }

\subsection{Experimental Results}
The comparative performance of our algorithm and other CNN/RNN-based models is detailed in Table \ref{tab:result}. The referenced models are cited alongside their respective names, with results derived from either open-source code or our own reimplementation. Test accuracy serves as the primary metric for comparison.

\review{As detailed in Table \ref{tab:result}, our algorithm achieves near-parity with leading models on datasets with limited vocabulary such as \textit{spam ham messages} and \textit{fake and real news}, trailing top performers by a mere 2\% while utilizing only 10\% of the parameters typically required. Conversely, its performance on datasets with a broader vocabulary range, notably \textit{Apple Twitter sentiment}, \textit{SST-2}, \textit{Agnews}, and \textit{Movie Review}, is less robust. The most significant challenges are observed with \textit{Agnews} and \textit{Movie Review}, where the intricate nature of the content, marked by extensive text samples and diverse vocabulary, notably diminishes the algorithm's efficacy. This trend underscores a pivotal limitation of our data compression approach, particularly the LZW algorithm's inability to effectively handle datasets characterized by low-repetition and high-diversity vocabularies. A more in-depth analysis of these constraints and their implications for our algorithm's performance across various dataset typologies is further explored in the following section.}

Beyond classification accuracy, our method's lightweight nature stands out, particularly its CPU compatibility. For instance, on the \textit{spam ham messages} dataset, a mere 23 atoms are selected from an initial set of 62,236 after LZW scanning, culminating in a 98.02\% accuracy rate. Remarkably, the entire process is completed within 67 seconds on a standard Intel Core i5-9600K CPU with 16GB RAM, underscoring our method's minimal computational demands.

Comparative details of computational and storage resources are presented in Table \ref{tab:parameters}. Notably, our model's parameter count and resource requirements are significantly lower than those of neural-network-based counterparts. Furthermore, our approach eliminates the need for pre-trained word vectors, streamlining the entire process. This efficiency is not compromised even when GPUs are considered, as our algorithm's computational demands are relatively modest. The LZW scanning process, a linear and hardware-accelerated operation, exemplifies this efficiency.

\begin{table*}[]
    \centering
    \caption{COMPARISONS OF THE APPROXIMATE NUMBER OF PARAMETERS AND EXECUTION TIME (THOUSAND : K) IN DIFFERENT MODELS }
    \begin{tabular}{c|c|c|c|c|c}
    \hline
        Models & The number of parameters & GPU support & Pre-trained word vectors & CPU training time(s)/epoch & GPU training time(s)/epoch \\
        \hline
        \hline
         CNN \cite{zhang2010discriminative} & 3,063K & Y & Y & 220 & 20   \\
        \hline
         RNN \cite{jang2020bi} & 7,646K & Y & Y  & 419  &  36  \\
        \hline
        Text GCN \cite{yao2019graph} & 21,741K & Y  & N  & N/A &  15 \\
        \hline
         Ours & 70K & N & N & 67 & N/A \\
         \hline
         
    \end{tabular}
    \label{tab:parameters}
\end{table*}

\begin{table*}[hbt]
    \centering
    \caption{Example Atoms with High Discriminative Power}
    \begin{tabular}{ |c||c|c|}
          \hline
     Datasets & Positive Atoms & Negative Atoms\\
      \hline
       \hline
     Spam ham messages & I go so my do at it here thing  & Number 0 to 9 ur to our call your mobile FREE claim !\\
    \hline
     Fake Real News & to in for say on of US with president & The To ID IN VIDEO Hi AT Of Video\\
     
    \hline
    Apple Twitter & Apple ! shope store with works Code workshops Hour Love & not my f*k one your hit sh*t need \\
    \hline
    
    SST2 &this that for usual what fake happy fun really exact & sad sh*t f*k super extreme so damn \\
    \hline

    \end{tabular}
    \label{tab:atoms}
\end{table*}

\section{Discussion}
\review{
In this section, we critically examine the broader implications of our experimental findings, offering a deeper understanding of the algorithm's performance, its inherent limitations, and the potential avenues for future enhancements.}
\subsection{Limitation of Data Compression Scheme}
\review{
Based on the experimental results discussed earlier, we now dive deeper into the performance nuances of our proposed LZW-based algorithm, with a specific focus on its application to the \textit{Agnews} and \textit{Movie Review} datasets. These datasets are notably challenging due to their rich vocabularies, a factor that critically influences the algorithm's efficacy.

At the core of our method, the Lempel-Ziv-Welch (LZW) algorithm is a foundational element, renowned for its proficiency in dictionary-based data compression. It operates by creating a new atom each time it encounters a novel string sequence during the sliding window process. While this approach is generally efficient, its performance is intrinsically tied to the dataset's repetition rate. In the case of \textit{Agnews} and \textit{Movie Review}, the broad spectrum of unique terms leads to a substantially lower repetition rate. This rarity of repeating sequences results in the generation of a vast number of atoms, culminating in a large and complex dictionary. Such a scenario starkly contrasts with binary streams, where the limited character set ensures higher repetition rates and simpler data structures.

This significant divergence in data complexity presents a pronounced challenge for our LZW-based algorithm, as evidenced by its underperformance on these particular datasets. Understanding these limitations not only sheds light on the algorithm's current constraints but also sets the stage for targeted enhancements in our future work, aiming to tailor our approach more effectively to the intricate nature of such text datasets.
}

\subsection{Insights into the Information-Plane Performance}
\review{
The information-plane performance of our algorithm, depicted in Fig. \ref{fig:info_result}, reveals critical insights. The observed space between our algorithm's trajectory and the optimal boundary, as computed by the information bottleneck principle, indicates potential for further optimization. Specifically, an improved algorithm might compress the original dataset more efficiently while preserving label-relevant information. However, as dictated by the inequality (\ref{Inequality}), any dictionary learning algorithm, due to its inherent constraints, will inevitably deviate from optimal performance. This phenomenon suggests that the gap between our algorithm and the theoretical optimum might not be as extensive as initially perceived.

Furthermore, this gap on the information plane correlates with the differences in performance between our algorithm and state-of-the-art models. In the first five datasets, where our algorithm exhibits competitive performance, the gap is noticeably smaller than in the last three datasets. This observation underscores the algorithm's relative efficacy across varied dataset types. Additionally, by dissecting the information trajectory with two vertical lines at its start and end points, we categorize the area under the information bottleneck boundary into regions \rom{1} and \rom{2}, as illustrated in Fig. \ref{fig:info_result}. Notably, datasets like fake and real news, Apple Twitter, and Agnews, with varying degrees of accuracy, demonstrate that the ratio of areas \rom{1} to \rom{2} is inversely proportional to accuracy. For instance, the fake and real news dataset, showcasing top-tier classification accuracy, presents a smaller area ratio compared to the Agnews dataset, where region \rom{1} predominates. Employing the Monte Carlo method for area estimation, we calculate the Information Plane Area Ratio (IPAR) for these benchmark datasets, yielding values of 0.3671, 0.3824, 0.5238, 0.5384, 2.5768, and 2.0434, respectively.}

\subsection{Exploring Interpretable Atoms}
\review{
The discriminative power inherent in our dictionary's atoms is a testament to the algorithm's refined update mechanism, as evidenced in the atom visualization presented in Table \ref{tab:atoms}. Our approach distinguishes itself by minimal data preprocessing—eschewing common practices like punctuation removal, stop-word filtering, and case normalization. This methodological choice is validated by the presence of discriminative atoms, including punctuation marks, which are typically disregarded in conventional word vectorization algorithms. By treating punctuation and other non-letter characters as standard elements, the LZW algorithm enhances the flexibility and comprehensiveness of the dictionary. For instance, subtle nuances like the distinction between lowercase "in" and uppercase "IN" significantly influence classification results, showcasing the algorithm's nuanced sensitivity to contextual variations.

Moreover, while the character-based version of our algorithm broadens the atom diversity, it also introduces interpretative complexity due to the prevalence of word substrings. This characteristic, despite potentially complicating our understanding of individual words, enriches the dictionary, offering a more nuanced representation of the textual data. Particularly in scenarios with limited training samples, this diversity proves advantageous, bolstering the algorithm's classification capabilities. The conceptual depth of each dictionary atom—encompassing its discriminative capacity, human interpretability, and semantic relationships—provides profound insights into the learning system's decision-making processes, highlighting the sophisticated nature of our approach and its potential for future enhancements.}

\section{Conclusion}

In this paper, we presented a novel supervised dictionary learning algorithm for text classification based on data compression. This is a two-stage algorithm where in the first stage, a dictionary is generated from the LZW algorithm. In the second stage, the dictionary is updated by selecting the atoms with high discriminative power to achieve better classification performance. The discriminative power is measured by evaluating the distribution of each atom over different classes. An information-theoretic analysis method is also presented to investigate the performance of our algorithms under this model using information bottleneck principles.

The effectiveness and accuracy of our algorithm are shown by the implementation on six benchmark text datasets. In addition, dictionary atoms with conceptual meanings are visualized and discussed for all datasets in terms of discriminative power, human interpretability and semantic relation. The information trajectories are plotted and compared with the optimal information bottleneck boundaries.

The performance on the information plane can be improved to reduce the gap between our algorithm and the optimal boundary. A combination of source coding and dictionary learning for text classification can be viewed as a promising research direction and its applications will be further investigated.  \review{We believe future research can focus on refining the algorithm for improved performance on complex, vocabulary-rich datasets, and optimizing the integration of the LZW scanning and dictionary update processes, targeting enhanced efficiency and accuracy.}

\bibliographystyle{IEEEtran}
\bibliography{main}
\vspace{12pt}
\end{document}